\documentclass[preprint,12pt]{elsarticle}
\usepackage{lipsum}
\makeatletter
\def\ps@pprintTitle{%
 \let\@oddhead\@empty
 \let\@evenhead\@empty
 \def\@oddfoot{}%
 \let\@evenfoot\@oddfoot}
\makeatother
\setcitestyle{authoryear,open={(},close={)}}

\usepackage{graphicx}
\usepackage{amssymb}
\usepackage{subcaption}

\usepackage{lineno}
\usepackage{booktabs}
\usepackage{xcolor}
\usepackage[font=small,labelfont=bf]{caption}
\graphicspath{ {./images/} }
\usepackage{tabularx}
\newcommand\setrow[1]{\gdef\rowmac{#1}#1\ignorespaces}
\newcommand\clearrow{\global\let\rowmac\relax}
\clearrow

\usepackage[symbol]{footmisc}
\begin{document}

\begin{frontmatter}


\title{A comprehensive solution to retrieval-based chatbot construction}



\author[1,2]{Kristen~Moore}
\author[3,4]{Shenjun Zhong}
\author[5]{Zhen He}
\author[]{Torsten Rudolf}
\author[]{Nils Fisher}
\author[5]{Brandon Victor}
\author[]{Neha Jindal}

\address[1]{CSIRO's Data61, Australia.\footnote{Corresponding author: kristen.moore@data61.csiro.au}}
\address[2]{Cyber Security CRC, Australia.}
\address[3]{Monash Biomedical Imaging, Monash University, Australia.}
\address[4]{National Imaging Facility, Australia.}
\address[5]{Department of Computer Science and Information Technology, La Trobe University, Australia.}

\begin{abstract}
In this paper we present the results of our experiments in training and deploying a self-supervised retrieval-based chatbot trained with contrastive learning for assisting customer support agents. In contrast to most existing  research papers in this area where the focus is on solving just one component of a deployable chatbot, we present an end-to-end set of solutions to take the reader from an unlabelled chatlogs to a deployed chatbot. This set of solutions includes creating a self-supervised dataset and a weakly labelled dataset from chatlogs, as well as a systematic approach to selecting a fixed list of canned responses. We present a hierarchical-based RNN architecture for the response selection model, chosen for its ability to cache intermediate utterance embeddings, which helped to meet deployment inference speed requirements. 
We compare the performance of this architecture across 3 different learning objectives: self-supervised contrastive learning, binary classification, and multi-class classification. 
We find that using a self-supervised contrastive learning model outperforms training the binary and multi-class classification models on a weakly labelled dataset. Our results validate that the self-supervised contrastive learning approach can be effectively used for a real-world chatbot scenario. 
\end{abstract}


\end{frontmatter}

\tableofcontents
\section{Introduction}
\label{S:1}

Customer service is one of the promising fields in which intelligent assistants can play a key role. This paper describes the response retrieval chatbot that we built and then deployed in early 2018 to assist (human) customer service agents at a telecommunications and technology company in Australia. At the time the model was built there were around 1.5K customer support agents, with projections to continue to grow the team to support the growing trend of customer interactions moving to chat-based channels.

In view of limitations of both the technology and the training dataset, to ensure quality control we chose to build a retrieval-based, internal facing chatbot. A retrieval based model ensures that responses can be vetted and modified by legal teams, and having the bot be internal facing means that the agent verifies the model's response before presenting it to the customer.

In order to deploy a real-life response retrieval chatbot starting from an unlabelled chatlog corpus, there were four problems we had to solve: 
\begin{enumerate}
    \item Develop a framework for the extraction and refinement of the canned response list.
    \item Create a training set from the raw chat logs of agent-customer interactions.
    \item Train a matching model for selecting the best canned response to propose.
    \item Learn when to not suggest a response, for cases where none of the canned responses are appropriate.
\end{enumerate}

To extract a high quality canned response list, we embedded the most frequently used agent utterances from the chat logs using a skip-thought model similar to that of \cite{SKIP}, then clustered the embedded utterances and selected a unique representative from each cluster. As a post-processing step, the wording of the canned responses was then refined by business and legal teams, so that several responses in the final list did not exist in the training data.

To create a dataset for training a canned response suggestion model, we used raw chatlogs consisting of 3.5M conversations and 110M utterances in total. Fully labelling such a dataset would require a person to inspect each agent response and decide if it is semantically equivalent to one of the canned responses. Since this would be prohibitively labor intensive, our first approach was to create a self-supervised dataset from the raw chat log corpus. This was done using the standard approach of treating human responses as positive samples and randomly sampling the negatives~\citep{UBUNTU}. In addition to this, a weakly supervised training set was created so that classification-based matching models could also be trained. Indeed, most state of the art matching models in the literature are supervised models, so anyone wishing to use one in a deployed chatbot would need to first create a labelled training dataset. As a prerequisite to creating a weakly labelled dataset, we first needed to curate the canned response list by inspecting the most frequent agent responses, as described above. Curating the list itself is challenging due to the many different ways agents can say the same thing. Then once the list is created, the task of finding the weak labels is even more challenging. Since it is too time consuming to manually inspect each agent response and determine if it semantically matches one of the canned responses, some automated method must be devised. To our knowledge, no prior work specifically addresses this problem.  To solve this problem, positive training examples were extracted from the raw chatlogs by comparing agent responses to the few hundred responses in the canned list using both ``exact" (regex-based) and ``fuzzy" (embedding-based) matching approaches. Negative training samples for the binary classifier were created using a skip-thought utterance similarity classifier. We call this dataset weakly labelled since both the fuzzy matching and the skip-thought utterance similarity classifier are by nature not precise, meaning the dataset can contain some false positive labels and false negative labels.

To create the matching model, we consider single-turn response selection models, where the entire conversation context is used for matching a reply from a canned list of responses. We take a hierarchical approach, using 3 modules for the model design: a word embedding, an utterance embedding, and a context embedding. This model architecture allows for caching of utterance embeddings, which enabled us to meet deployment inference speed requirements. We then measure the performance of this RNN-based architecture on three different learning tasks: contrastive learning, binary classification, and multi-class classification. Since we initially had no labelled data, our deployed real life chatbot used the self-supervised contrastive learning model that learns to match a context embedding with the desired agent utterance embedding from the raw data. Then later, once a weakly labelled dataset was created, we trained supervised binary and multi-class classification models so that we could compare their performance with the self-supervised approach. 

Finally, to address the problem of when to make responses, a threshold was set for the minimum required model confidence, to prevent the model from making recommendations when there weren't any suitable canned responses. 

In contrast to most existing  research papers in this area where the focus is on solving  just one of the above problems, we had to solve all four of them in order to produce a functioning system for deployment. The main research contribution of this paper is the development of an end-to-end framework for building and deploying a real-life response retrieval chatbot, starting from an unlabelled corpus of chatlogs. More specifically, we make the following key contributions: 
\begin{enumerate}
    \item We present a hierarchical, self-supervised response matching model that uses contrastive learning to select canned responses, and validate this approach on a real-life chatbot, ie. train on hundreds of thousands of distinct, unlabelled targets, and deploy it on only a few hundred ``canned response" targets, which may not exist in the training set. 
    \item We present an alternative, supervised approach that includes constructing a weakly labelled dataset from the raw chatlogs, and then training a binary or multi-class response selection classification model. To the best of our knowledge, this is the first multi-class classification response matching model in the literature.
    \item We present an approach for extracting a canned response list from an unlabelled chatlog corpus, and developing a method for detecting when no suitable response from the canned list can be suggested.
    \item We present a deployment architecture for serving our chatbot model on an enterprise customer support messaging platform. 
\end{enumerate}

\section{Related Work}\label{Related Work}

\paragraph{Response retrieval models}
In this work we present a single-turn response selection model, where the entire conversation context is used for matching a reply from a canned list of responses. 

\noindent A fundamental challenge in response selection is measuring the degree of matching between a conversation context and a response candidate. The focus of most existing research to date has been on constructing a matching model through sophisticated neural architectures. Early work on the problem by \cite{UBUNTU} used a Dual Encoder LSTM approach, which individually encodes the concatenated context and a response candidate with LSTMs, and then calculates a matching score based on the final states of the two LSTMs. Such an approach can be considered a word sequence model, as it models the word as a unit and views the context and response as sequences of words.
\cite{Zhou} extended the word sequence model approach with their multi-view model, which integrates information from both the word level and the utterance level. \cite{Chaudhuri} extended the work of Lowe et al. by adding two attention mechanisms in addition to incorporating external knowledge (eg. Ubuntu manual descriptions) via domain knowledge embeddings.

To better capture the relevance of the dialogue context and the response, multi-turn matching systems have recently been proposed, which match the candidate response with every utterance of the context ~\citep{SMN, Zhang}. \cite{SMN} present a hierarchy-based approach which matches a response with each utterance in the context, and then distills and fuses the information using a convolutional neural network approach. \cite{Zhang} introduces a turns-aware deep utterance aggregation model that leverages self-matching attention to form a fine-grained context representation.

Following the success of a self-attention-based model on various NLP tasks~\citep{Tran}; self-attention was also  adapted for the response selection task~\citep{DAM, Tao, Yuan}.   \cite{DAM} used self-attention and cross-attention to construct multi-grained representations, and compare their similarity at each granularity. \cite{Yuan} proposed a multi-hop selector to filter only relevant utterances to a response. \cite{Whang} proposed a BERT based post-training method for a multi-turn conversational system.

All of the aforementioned works have used recall in position $k$ from 10 candidates, denoted $R_{10}@k$, as a key metric for evaluating their models (defined in detail in section \ref{evaluation-metrics}). This means they get the model to select the top $k$ responses from $10$ candidate responses. However, when building goal-oriented dialog systems, the set of possible responses is usually much larger. Addressing this limitation was a goal of the first track of the Dialog System Technology Challenge 7 (DSTC7) \citep{DSTC7}, which required participants to select the correct next response from 100 candidates (measured by the $R_{100}@k$ metric) on the Ubuntu and Student-advisor datasets. The best performing model on subtasks 1, 4 and 5 of the challenge was the Enhanced Sequential Inference Model (ESIM) of \cite{CHEN2020101072}, based on the model of \cite{chen-etal-2017-enhanced} that was originally developed for natural language inference.

Selecting the right response from 10 or 100 candidates, whose negative examples are sampled from a much larger list of all possible responses, is still a different problem to what we encountered in the chatbot deployment scenario. In a real life chatbot, the matching model would select from the same full list of canned responses at every turn. Another key difference is that we often do not have an exact match between the agent response and any of the canned responses in the canned list, however the agent's response may have the same meaning as one of the responses in the canned list. This deployment response matching problem is the one we consider in this paper. Furthermore, we had to solve a more holistic problem that included extracting the canned list from the raw chat logs and creating a training dataset in addition to designing the matching model that selects a candidate response from the canned response list, and deciding whether or not to propose a response suggestion at each turn based on the input chat context.\\

\paragraph{Canned list extraction} This problem arises in the chatbot deployment setting, and to the best of our knowledge, has not been discussed in the literature.

\paragraph{Creation of a labelled dataset} 
There are a number of different problems that have been studied in the literature around crafting and improving the dataset for response retrieval chatbots. The standard approach for response retrieval models is to formulate the task as a classification problem and construct a self-supervised training dataset by treating human responses as positive examples and then randomly sampling negative responses. However, such a training set might contain lots of false negatives (samples with the same semantic meaning to the target response), as well as true negatives that are very easy to distinguish from true positives because they are far from the semantics of the contexts. As a result, models with advanced architectures have been found to do well when trained on such datasets with randomly sampled data (eg. Ubuntu), but only reach sub-optimal performance on human labeled data (eg. Douban)~\citep{Wu3}. Optimising the training dataset for the task of learning can therefore be seen as another aspect of the response matching challenge. 

In the binary classification response matching model setting, \cite{DAM2} studied how to optimise the sampling of negative examples and proposed four sampling strategies. When compared with the widely used random sampling strategy, they found that two of their strategies lead to performance drop, but the other two brought consistent improvement to the performance of all the 3 models they tested.

In the contrastive learning literature, negative sampling is a well-studied problem. \cite{Samp} show that when negative examples are too close/hard, the gradient has high variance and it has low signal to noise ratio. This is the case for hard negative mining, which when applied to the triplet loss, can lead to noisy gradients that cannot effectively push two examples apart, and consequently a collapsed model. On the other hand, random samples are often too far apart to yield a good signal, and thus lead to no progress for learning. Semi-hard negative mining \citep{SHT} on the other hand finds the narrow set in between, yielding a violating example that is fairly hard but not too hard. To the best of our knowledge, the only work in the literature that considers contrastive learning in response retrieval chatbots is the paper of \cite{Wu3}. They took existing binary classification matching models from the literature and replaced their binary cross entropy loss with a max-margin loss, which used a sequence-to-sequence architecture coupled with a search engine to achieve sophisticated negative sampling.

In this work we study a different problem, which arises in real world chatbot deployment scenarios. Suppose we are given the canned response list that is to be used for the deployed chatbot, which may have been handcrafted and/or refined by business and legal teams, and thus include responses that are not present in the raw chatlogs. How can we then optimise the creation of a supervised learning training dataset for binary and multi-class response matching models, given the canned response list and the raw chatlog?

\paragraph{Deciding when a response should be given} The DSTC7 challenge included a subtask that required the dialogue system to be able to recognize cases where no suitable response is available in the set of candidate responses. For subtask 4 of the challenge, the candidate set of responses contained 100 responses that may not include a correct response, and the system must respond with a `None' response when no correct response is found. Since a labelled training set was provided for the subtask, several participants chose to apply their response selection model without making any architectural changes to explicitly address whether or not a response should be given. \cite{CHEN2020101072} used a threshold on the model confidence to decide when to predict that the candidate pool did not contain the correct next utterance. \cite{BOUSSAHA2020101080} took an alternative approach and applied a classifier on top of the response selection model to decide whether or not a correct response exists among the candidate responses.

In our real world chatbot deployment setting, we didn't have the ground truth labels, and thus could not formulate this as a supervised learning problem. To the best of our knowledge, how to solve this in the unsupervised setting has not been discussed in the literature.

\section{Proposed end-to-end framework}\label{end-to-end}
\noindent Starting with only the raw logs from historical customer-agent chats, our aim for the pilot phase of the chatbot was the following:
\begin{enumerate}
\item Train a predictive model for the next agent response out of a list of canned responses.
\item Finding a good list of canned responses.
\end{enumerate}

\noindent Since we had no labelled training data, our initial focus was on self-supervised models that learned from the raw chatlog data. The self-supervised model described in section \ref{max-margin LSTM} was chosen for deployment, and a global threshold was set for the minimum required model prediction confidence, so that the model would recommend responses roughly 70\% of the time. This was generous given the small size of the canned response list (initially 153 responses, then later extended to 290 responses). The chatbot was deployed in a small widget within the agent’s chat UI on LivePerson. Whilst a chat was taking place, the widget was automatically populated with the top 3 model responses. On top of that, frequently used `actions’ were also displayed, such as greetings and customer authentication at the beginning of a chat, prompting an inactive customer later in the chat, and resolution confirmation towards the end.

\begin{figure}[h]
\includegraphics[width=13.6cm]{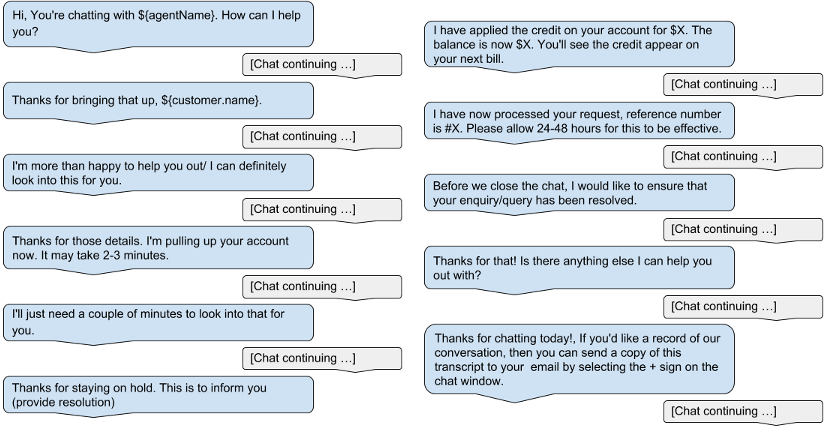}
\caption{Example of canned response usage. The blue bubbles contain canned responses, and the grey bubbles represent customer responses as well as agent responses in cases when a suitable canned response didn't exist.}\label{example canned}
\end{figure}

Deployment of the chatbot started in early 2018 with a pilot phase of 30 customer service agents, and grew from there to over 1,000 agents. Weekly usage stats from August 2019 show the chatbot being used in roughly 50K chats per week, which represents 65\% of all chats. Overall 18\% of all agent responses came from selecting a response recommended by the model, with some chat channels like ``prepaid mobile" having an average model usage rate above 30\% (as some of the canned responses were targeted to that channel). A typical use of the chatbot model suggestions is shown in figure \ref{example canned}, giving an indication of the types of contexts for which canned responses were used.

\subsection{Dataset}
The chat log dataset consisted of roughly 3.5 million conversations that were rigorously processed to remove all personally identifiable information (PII), including (but not limited to) details relating to names, addresses, email address, and date of birth. On top of that, all digits were converted to zeros, and the resulting cleaned dataset as well as the PII removal process was approved by security architecture before any work on the project began. High level statistics on the dataset are provided in Table \ref{tab:data-stats}.

  \begin{table}[h!]
  \begin{center}
    \begin{tabular}{|l | c |}
      \toprule 
      \# dialogues & 3.5M \\
      \midrule 
      \# utterances (in total)& 110M\\
      \midrule 
      Avg. \# words per dialogue & 464 \\
      \midrule 
      Avg. \# words per agent utterance & 26.7\\
       \midrule 
      Avg. \# words per customer utterance  & 11.6\\
      \bottomrule 
    \end{tabular}
    \caption{Dataset statistics.}\label{tab:data-stats}
  \end{center}
\end{table}

\noindent There are a few noteworthy observations about the raw dataset. Firstly, the solution to the customer's problem often isn't captured in the dialogue. Eg. The agent resolves the problem by making updates to a database, and then replies to the customer saying only that they have fixed the problem. Secondly, conversations are full of pleasantries and customer authentication steps that aren't relevant to the customer's problem. So even though the human agent is solving customer problems through interactions with multiple systems, the chatbot is trained only on the text the agent writes to the customer. Some data challenges include lots of typos, slang, and bad grammar, as well as dialogues where the agent doesn't resolve the customer’s problem, which we had to decide whether to keep or remove from the dataset.

\subsection{Construction of the canned response list}

A few challenges had to be taken into account when creating the initial canned response list. Firstly, the canned responses had to seem natural, as the customer service agents had highlighted this as an important issue driving their willingness to use the suggested responses. Secondly, restricting our model to only selecting from a list of canned responses ultimately limits the power of our model, so it is important to have representative responses from all the most important/frequent agent intents. Thirdly, we had no labelled data, only historic chats, so for this reason our focus was initially limited to self-supervised models.

Since the agents tended to re-use utterances from both centralised and non-standard lists, this meant there were some utterances in the raw chat logs that were identical/very similar. Our high level idea was to semantically cluster the most frequently used utterances, and choose a unique representative from each cluster. To create the initial set of canned responses, we used the raw chatlogs to train an utterance embedding model based on the skip-thought model of \cite{SKIP}, with the slight difference that during training we used the current agent utterance to predict the 2 previous and 2 following utterances. We then used the resulting embedding to embed the most frequent 10K agent utterances (where the top 10K represented approximately 1/3 of all the unique utterances). K-means clustering was then performed on top of this embedding to separate the responses into 200 clusters. Unique representative intents were then manually chosen from each cluster. After a further round of refinement and approval from the legal team, this produced an initial canned list of 153 responses.

An utterance similarity classification model (described in \ref{utterance_sim}) was used later to find frequent agent utterances from chat logs that were dissimilar from all existing canned responses. This brought the canned response list up to 290 unique responses.

\subsection{Response retrieval model architecture and training}\label{model-comparisons}
As a first approach, the contrastive learning RNN model described below was pursued to see if a model could be learned directly from the raw, unlabelled data. This model was deployed to the live chatbot at the beginning of 2018. 

Later on, once we had collected several months worth of labelled data from the deployed chatbot, a labelled dataset was constructed (see section \ref{Supervised-dataset}). This enabled us to train both binary and multi-class classification matching models (described in section \ref{supervised models}), so that we could compare the performance of self-supervised with supervised learning approaches.

At inference time, our contrastive learning model selects the closest canned response to the given context, with regards to cosine similarity. On the other hand, a binary classifier model compares the context with every canned response and picks the highest confidence one, and a multi-class classifier outputs the ID of the response with the highest confidence.

\paragraph{Training dataset creation}
The training dataset for the contrastive learning model was created from the raw chat log corpus by treating human responses as positive samples, then all other targets within the batch were considered negative responses. Detailed instructions on this training set generation are given in \ref{training-set}. 

\paragraph{Contrastive learning model}\label{max-margin LSTM}
A self-supervised learning approach was taken, using $<$context, target$>$ pairs created from the raw, unlabelled dataset. The philosophy of our approach is to embed the chat context and the canned responses into the same space and do similarity matching to find which canned responses to suggest. To facilitate this, a max-margin ranking/contrastive loss function is used during training to pull the context and the true target together while pushing the context and negative targets away from each other.  Our formulation of the max-margin loss $L$ in equation (\ref{N-pair-loss}) applies the standard triplet ranking loss function to all $O(N^{2})$ triplets of the form $\left(c, p, n \right)$ in the batch,

\begin{equation}\label{N-pair-loss}
L =  \frac{1}{N}\sum\limits_{i=1}^{N}\sum\limits_{n=1}^{N-1} max(0, m + D(c_i, p_i) - D(c_i, n_n))
\end{equation}
where $c$ denotes the context, $p$ is the positive target, $n$ is the negative target, and $N$ is the batch size. The loss function (\ref{N-pair-loss}) leverages a margin \textit{m}, whereby the \textit{max} function together with the margin, \textit{m}, make sure that any negative targets that are at a distance greater than \textit{m} from the context do not contribute to the ranking loss. We use cosine distance as the distance function $D$.\\

Our model architecture is based on the Dual Encoder LSTM model~\citep{UBUNTU}, which used a Siamese network consisting of two RNNs with tied weights to produce the embeddings for the context and response. As input, their model takes both the context and response, which are each flattened to a sequence of words.

\begin{figure}[h]
\includegraphics[width=14cm]{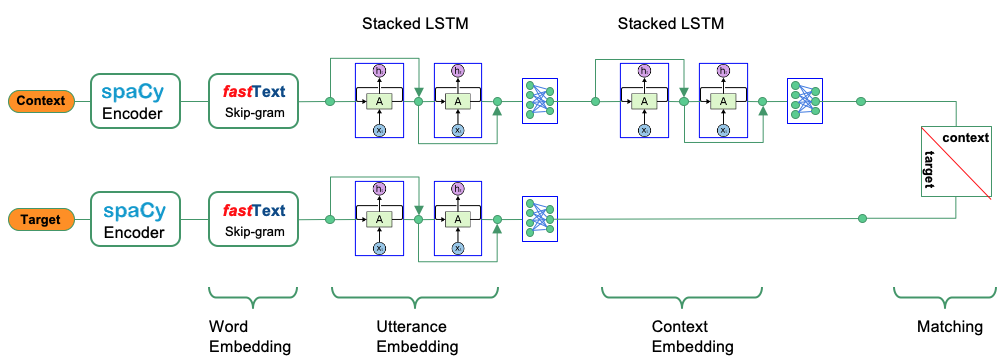} 
\caption{The hierarchical architecture of the contrastLSTM model.}\label{max-margin-architecture}
\end{figure}

Our model, \textit{contrastLSTM}, extends upon this approach by introducing a hierarchical architecture shown in Figure \ref{max-margin-architecture}, which features 3 modules: a word embedding, an utterance embedding, and a context embedding. Firstly, the words from the context utterances are projected into word embedding vectors, then an utterance embedding module encodes each utterance into a single vector, and lastly the context embedding module encodes the entire conversation context into a single vector (or into memory slots, in the case of memory networks). In parallel, the target utterances are also embedded using the word and utterance embedding modules. Finally a linear layer is applied to the output of both the target embedding and the context embedding, projecting them both to a vector of the same size. Shared weights were used across the context and target utterance embedding modules, though we found that it did not make much difference whether or not weights were shared.

Compared to the approach of \cite{UBUNTU}, our hierarchical architecture has the advantage of learning a multi-grained representation of each dialogue, at the word, utterance, and context level. Furthermore the input sequences per RNN are much shorter than in \cite{UBUNTU}, where the length of the input sequence for the context is equal to the number of words in the context. Other differences in our approach include the use of layer norm, as well as a different loss function - our training approach employs a max-margin loss function whereas cross entropy is used in \cite{UBUNTU}.

For the utterance embedding module we tried several approaches, including BOW, GRU and (bi)LSTM. For the context embedding module we tried GRU, (bi)LSTM and memory networks (Bordes et. al. 2017). The best results were obtained using stacked LSTMs for both the utterance and context embedding modules. 

\paragraph{Training process and hyperparameters}
A batch size of 128 was used, which means the training objective was to choose the correct response from a list including the correct target and 127 negative targets. A margin $m$ of 0.0001 was used for the max-margin loss. The utterance and context embedding modules each consisted of 2 layers of LSTM with residual connections. Layer norm was applied in the utterance and target embedding modules, and a dropout keep-probability of 0.5 was applied before the linear output layer of the target and context embeddings. Overfitting was not a problem we encountered with this architecture. We investigated using different embeddings for different utterance types i.e. customer Vs agent utterances, (which was one key focus of \cite{Wang2}), however no performance improvement was observed, so this idea was abandoned.  


\subsection{Deciding when to suggest responses}\label{when-to-suggest}
The pilot version of our deployed chatbot had a canned list of only 153 responses, therefore for the majority of turns in the conversation there wasn't an appropriate canned response for the agent to use. So to reduce the amount of wasted time reading irrelevant response suggestions, we needed a mechanism for deciding whether or not to suggest responses at each turn in the conversation.

A simple heuristic for deciding when to suggest responses is to set a threshold for the minimum required model confidence. The decision was made to prioritise recall over precision for the pilot phase of the chatbot, and a threshold was accordingly chosen, which resulted in the chatbot recommending responses roughly 70\% of the time.

The model confidence anlaysis relevant to this problem is presented in section \ref{Model confidence analysis}.

\subsection{Model serving architecture and optimization}
\begin{figure}[h]
\includegraphics[width=13.5cm]{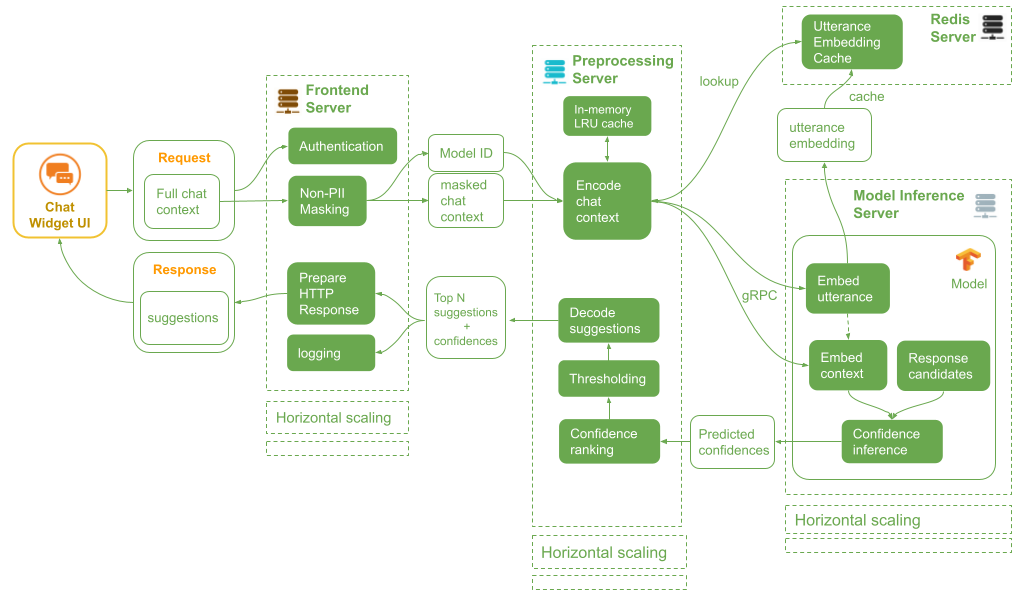} 
\caption{Chatbot deployment architecture.}\label{deployment-architecture}
\end{figure}

Our chatbot model server was deployed on AWS with the model serving architecture shown in Figure \ref{deployment-architecture}. 

The system runs multiple servers running python or Scala and React JS on the client-side. Each of the servers is designed so that it can be horizontally scaled and at least 2 of each were typically deployed for redundancy. This setup was created on AWS infrastructure using Terraform scripts for reproducible infrastructure as code.

Each time a conversation has a new utterance we request a new set of predictions to display to the agent. To produce this result, the context conversation goes through multiple processing stages described below, before being returned as predictions. Notable side-effects from each request are caching of embeddings and logging of usage statistics for future training and analysis.

\paragraph{Chat Widget UI}
The LivePerson chat widget runs React JS in an Iframe to display and request the chat suggestions and collect usage information. It also performs initial PII masking (the removal/obfuscation of Personal Identifiable Information) on the client side. Both requests and usage information (if a suggestion is used) are sent to the front-end  Server.

\paragraph{Front-end Server}
The front-end server consists of a Scala Akka HTTP server running API endpoints and serving the static code for the widget. It handles authentication and receives the full chat context from the widget. It then performs additional non-PII masking (for example conversion of digits to non-specific masking strings, 94567 -> XXXXX) and passes the chat context on to the data preprocessing server. Responses from the data preprocessing server then get passed back to the React front end. In addition to this, the front-end server logs (using DynamoDB) the model suggestions and the agent's suggestion usage statistics.

\paragraph{Data Preprocessing Server} 
The data preprocessing server is a Python application that is fronted by an aiohttp HTTP server, which streamlines utterance embedding and context similarity requests over gRPC to the TensorFlow Serving model inference server. Individual utterances are preprocessed using spaCy and then sent for embedding. When all utterance embeddings are available they are sent to the model inference server which responds with a top 10 canned response suggestions. Those 10 are filtered against recent 3 agent utterances to avoid repetition. It then compares model confidence of the top response against a minimum confidence threshold. If the top response is above the threshold, then the top 3 responses get passed back up to the chatbot server in JSON format.

 The requests to the model inference server are broken up into the 2 tiers to allow us to cache the utterance encodings and embeddings, which otherwise would be a large unnecessary preprocessing and embedding load each time utterances are processed. However this comes with the trade-off of doing two requests (firstly for the missing utterances and then for the entire context) where a single end-to-end request would work. The load reduction on the TensorFlow Serving infrastructure is worth the overhead of two requests unless none of the embeddings are cached. The first layer is a cross server Redis cache keyed on unprocessed text with values being final utterance embeddings. A second layer is a per server in memory python LRU cache with unprocessed text keys and encoded (but not embedded) utterances as values, which reduces the impact of Redis cache misses. The cross server Redis cache allows the number of preprocessing servers to be scaled up. These caches assist primarily with not re-calculating encodings/embeddings for ongoing conversations, where only a single utterance each per agent and customer is appended (and once the utterance limit of context is reached, one utterance is removed from the start) between each request. Ideally the Redis cache is at least large enough to cache at least all concurrent ongoing context utterances. There is also benefit from cross context reuse eg. agent utterances (canned responses or otherwise) or PII masked customer utterances.
 
\paragraph{Model Inference Server}
This is just a basic TensorFlow Serving deployment which listens over gRPC. The server runs an exported model described in section \ref{model-comparisons}, and hooks into the start of the context chain for non-cached utterances and post the first stacked LSTM for the recommendation requests visible in Figure \ref{max-margin-architecture}. For utterance embeddings it expects input utterances based on the universal vocabulary, and returns the embedding values after the first stacked LSTM. For recommendation requests it expects a set of utterance embeddings representing the context and then calculates the confidence of each canned response. The model confidences for the canned responses are then sent back to the data preprocessing server.

\subsection{Weakly labelled dataset generation (optional)}\label{Supervised-dataset}
The main benefit of a self-supervised matching model, like contrastLSTM, is that it does not require a labelled training dataset. Therefore this section should not be considered part of the end-to-end deployment of our contrastLSTM chatbot, and why we have put it at the end of section \ref{end-to-end}. However, since all state of the art matching models in the literature are supervised models, anyone wishing to use one of those in a deployed chatbot would need to first create a labelled training dataset. For us, it was necessary to create a labelled dataset to be able to compare the performance of our contrastive learning model to supervised matching models.

The contrastive learning approach does not train the model to directly predict when and which canned response to use. Since this potentially limits its accuracy when used for predicting the correct canned response, and also whether any canned response can be used, we wanted to try a multi-class classifier that was directly trained using canned responses as targets. In order to do this, we had to create a weakly supervised training dataset of $<$context, target$>$ pairs where the targets consisted only of canned responses. We call this the `positive' dataset.

\begin{figure}[h]
\includegraphics[width=8cm]{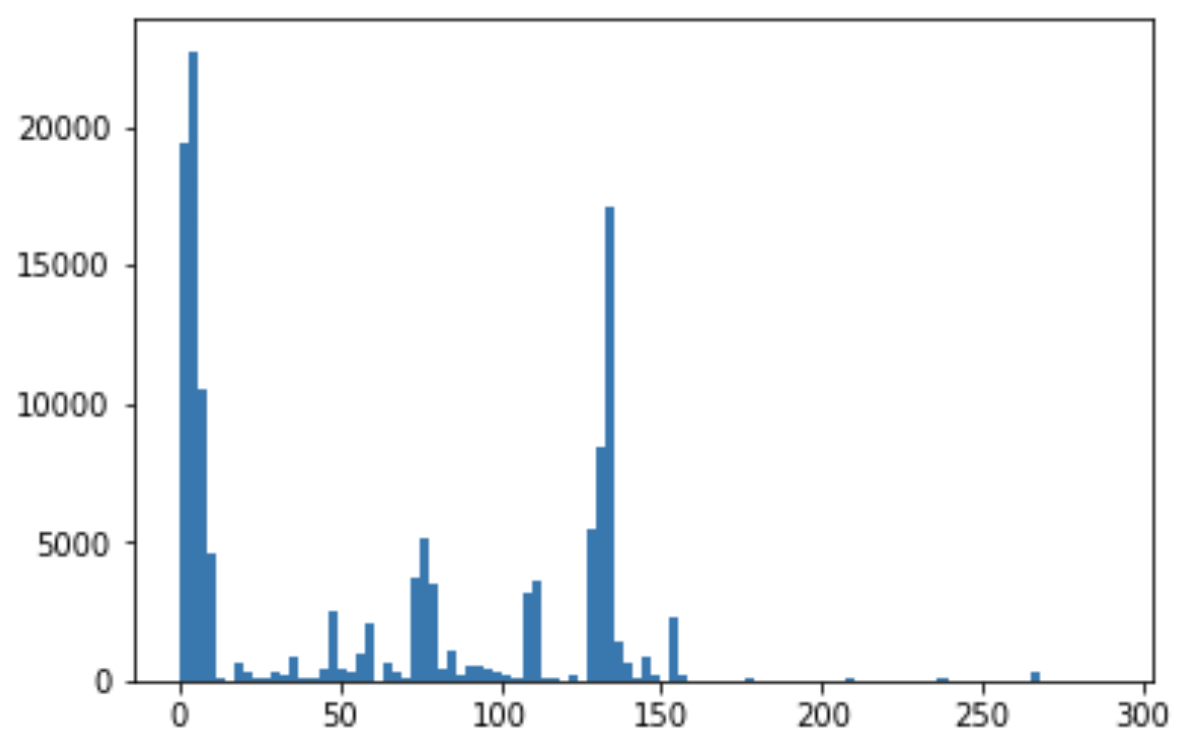} 
\caption{Distribution of counts across the 290 canned responses in the positive dataset.}\label{positive-distribution}
\end{figure}

\paragraph{Positive dataset }
When we came to the task of creating the labelled dataset, the chatbot had already been deployed with the contrastive learning model from section \ref{max-margin LSTM} for a number of months. This meant we had roughly 100K high quality positive samples from the usage logs of the deployed chatbot, where the agent had used the suggested canned response. Roughly 8K of these positive samples were set aside as a test set, and another 8K allocated to a validation set. To augment this dataset, we used both exact-matching and semantic fuzzy-matching approaches to extract further positive pairs from 2.8M raw historical chatlogs, as well as from the usage logs from the deployed model. 
\begin{enumerate} 
\item Around 1.95M absolutely matched positive data pairs were found using regular expression string matching. These are high quality samples.
\item Around 1.86M fuzzy matched positive data pairs were found using the similarity model of \ref{utterance_sim} to match the agent utterance with a 'similar' canned response. These are of lower quality than the absolutely-matched samples. Roughly 220K of these were set aside as a test set.
\end{enumerate}
This resulted in around 3.9M positive pairs to use for training the classification models. However, as shown in figure \ref{positive-distribution}, the distribution of counts for the responses is very skewed.  The highly skewed distribution makes training an accurate model particularly difficult since there is not have enough diverse examples of the minor classes to train a classifier that can effectively generalize to the test set. 

\paragraph{Negative dataset}
\noindent The creation of the positive dataset provided the opportunity to train the LSTM binary classifier on a weakly labelled dataset of canned responses. However, to be able to do this, we also had to create a negative dataset. A negative dataset consists of pairs of contexts with corresponding incorrect responses.   This is a very challenging problem for several reasons. Firstly the number of potential negative examples is massive. So it is hard to create a negative dataset that covers all potential situations likely to be encountered during testing. Secondly, there are a lot of conversation contexts where we do not know for sure which, if any, of the canned responses can be used. We had to resort to using our utterance similarity model to select the negative responses. This in turn can produce false negatives. 

There is a lot of scope for creating negative pairs. There are two types of contexts that need to be covered by the negative dataset. Firstly, we wanted good coverage from the contexts of the positive dataset, so we took contexts from the positive data set and converted them into negative examples by attaching an inappropriate target from the canned response list. Secondly, we wanted good coverage of all the contexts from the raw chatlog for which no appropriate canned response exists (according to our similarity model). Such contexts can be combined with any canned response to make a negative $<$context, target$>$ pair. The utterance similarity model was used to generate the negative pairs for both of the above approaches. Lastly, we added examples from the model usage log where the agent didn't use the suggested canned responses, adopting the heuristic that the suggested response wasn't used because it wasn't an appropriate response for that conversation context. However, this heuristic has the potential to include false negatives in the dataset since agents may never have been offered the correct canned response by the deployed model, or the agent chose not to use the suggested canned responses even when it was correct.

\section{Experimental setup}
To validate self-supervised contrastive learning as a viable approach for training a matching model, we compare our LSTM hierarchical embedding architecture across 2 additional learning objectives:  binary classification and multi-class classification. We also compare the performance of constrastLSTM with the Deep Attention Matching (DAM) network.
\subsection{Supervised learning baselines}\label{supervised models}
\begin{figure}[h]
\includegraphics[width=12cm]{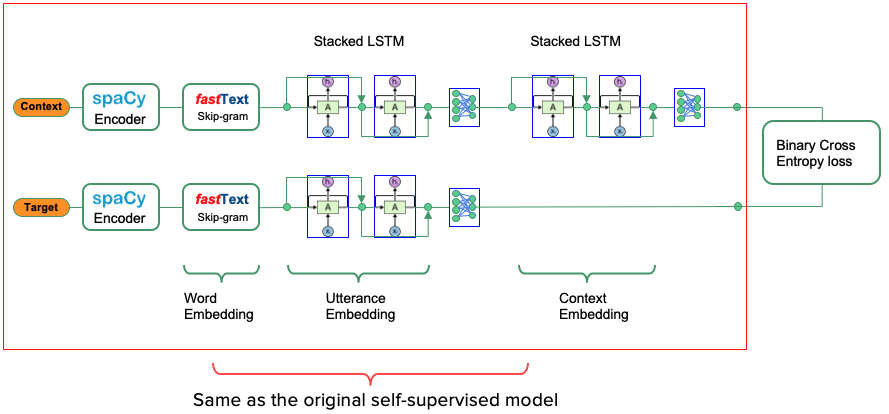} 
\caption{The binary classifier uses the same hierarchical architecture as contrastLSTM, but with binary cross entropy loss instead of the contrastive max-margin loss.}\label{binary-classifier}
\end{figure}

The following 3 supervised learning baselines are used in our experiments.
\paragraph{i) Binary classifier}\label{Binary-classifier} 
The binary classifier supervised learning model architecture is shown in Figure \ref{binary-classifier}. It is just a reformulation of contrastLSTM, with stacked LSTMs forming the utterance and context embeddings, however now a sigmoid activation function is applied to the concatenated context and target embeddings, and binary cross-entropy loss is used for training. 

\paragraph{ii) Multi-class classifier}

\begin{figure}[h]
\includegraphics[width=12cm]{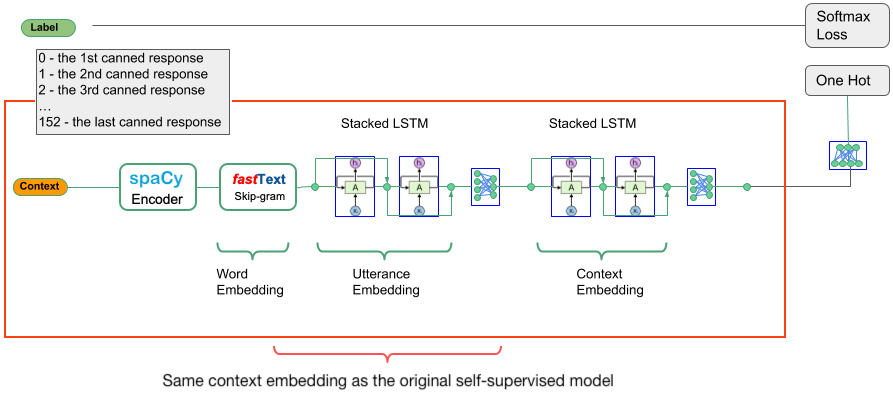} 
\caption{Multi-class classifier architecture.}\label{multi-classifier}
\end{figure}

Instead of embedding both the context and the target, the multi-class classification supervised learning model shown in Figure \ref{multi-classifier} takes only the context as input and then outputs the ID of the correct canned response (or top 3 responses). Again, the model architecture uses stacked LSTMs for the utterance and context embeddings, however now the embedded context is passed to 3-linear layers. The last linear layer outputs an N-dimension tensor, (where N is the number of canned responses) and then the softmax activation function is applied to indicate the confidence of picking each canned response. 

A range of activation functions were trialled for the first 2 linear layers, and both SGD and the Adam optimizer with learning rate decay were tested, however only the SGD optimizer together with a tanh activation function worked well for this dataset.

\paragraph{iii) DAM model}
Following the success of the DAM network of \cite{DAM} on the Ubuntu Dialogue corpus, we decided to test it on our unlabelled chatlog dataset. Further discussion of this model can be found in \ref{DAM classification model}.

\subsection{Evaluation metrics and dataset}
It is difficult to fairly compare the performance of the 4 models since they are trained using different datasets, and in essence solve different problems. The contrastLSTM model of section \ref{max-margin LSTM} learns a similarity function that measures the distance between samples. On the other hand, the other models are either single-class classification problems, in the case of both the LSTM binary classifier and the DAM network, or a multi-class classification problem, in the case of the LSTM multi-class classifier. 
 
In light of this, the fairest way to compare the 4 models is to replicate the chatbot deployment scenario, where the model recommends responses from the list of 290 canned responses. 

As mentioned above, in this deployment setting, the max-margin LSTM selects the closest canned response to the given context, with regards to cosine similarity. On the other hand, the DAM and LSTM binary classifiers each compare the context with every canned response and pick the highest confidence one, and the multi-class LSTM outputs the ID of the response with the highest confidence.

\paragraph{Evaluation dataset}
We use the test dataset of 8K high quality positive samples from the usage logs of the deployed chatbot model (as described in section \ref{Supervised-dataset}). 

\paragraph{Evaluation metrics}\label{evaluation-metrics}
To measure the performance of each of the models, we use the standard Recall@k evaluation metrics as in previous works \citep{UBUNTU}, (denoted $R_n@k$ below). Given a conversation context, each model is asked to select the $k$ most likely responses from $n$ available candidates, and the model response is considered correct if the true response is among these ``Top-$k$" candidates.  Since our deployed chatbot showed the customer support agents the top 3 responses, the $R_n@3$ or ``Top-3" accuracy was an important metric for us. 

\section{Results}\label{Experiments}
\subsection{Comparison of constrastLSTM with DAM} 
The following two tests were carried out to evaluate the performance of the contrastLSTM using the $R_n@1$, $R_n@3$ and $R_n@10$ recall metrics, which are defined in section \ref{evaluation-metrics}.
\begin{enumerate}
\item Measure the Top-1, Top-3 and Top-10 recall rates from the full list of 290 canned responses (including the true response). This test replicates the deployment scenario, where the model recommends responses from the list of 290 canned responses.
\item  Measure the top-1, top-3 and top-10 recall rates from a list of 128 random agent responses from the raw chatlogs (including the true agent response). This test replicates training where the true response must be chosen from the batch size of 128 responses. 
\end{enumerate}
 For each test we randomly sampled 5,000 (context, response) pairs, and then averaged the results over 10 repetitions. These (context, response) pairs were taken from the test dataset of the 100k high quality positive samples from the usage logs of the deployed chatbot model (further details on this positive dataset is found in section \ref{Supervised-dataset}). 
 
The results are shown in Table \ref{LSTMvDAM}. The results for the DAM model with \textit{n} attention layers are denoted by DAM-\textit{n}, and the results for the contrastive learning LSTM model are denoted by contrastLSTM.

\begin{table}[h!]\
 \begin{center}
  \begin{tabular}{>{\rowmac}c|>{\rowmac}c >{\rowmac}c >{\rowmac}c >{\rowmac}c <{\clearrow}}
   \hline 
     & Avg-Pos. 290 & R$_{290}$@10 & R$_{290}$@3 & R$_{290}$@1\\
   \hline 
   DAM-1& 18.84 & 0.626 & 0.407 & -  \\
   DAM-2& 21.77 & 0.587 & 0.520 & - \\
   DAM-3& 18.71 & 0.644 & 0.431 & - \\
   DAM-4& 19.16 & 0.641 & 0.435 & - \\
   DAM-5& 18.81 & 0.639 & 0.427 & - \\
   \setrow{\bfseries}contrastLSTM & 2.88 & 0.950 & 0.847 & 0.681 \\
   & & & & \\
   \hline 
   & Avg-Pos. 128 & R$_{128}$@10 & R$_{128}$@3 & R$_{128}$@1 \\
   \hline 
   DAM-1& 8.82 & 0.764 & 0.514 & -  \\
   DAM-2& 8.68 & 0.770 & 0.520 & - \\
   DAM-3& 8.77 &  0.764 & 0.522 & - \\
   DAM-4& 8.53 & 0.776 & 0.528 & - \\
   DAM-5& 8.45 & 0.778 & 0.533 & - \\
   \setrow{\bfseries}contrastLSTM& 2.11 & 0.977 & 0.876 & 0.698 \\
   \hline 
   \end{tabular}
  \end{center}
  \caption{Comparison of our contrastLSTM model Vs DAM on various ``Top-k" recall metrics $R_n@k$.}\label{LSTMvDAM}
\end{table}

Based on these metrics, the accuracy of the DAM models is much worse than the contrastLSTM model. It is possible that contrastLSTM is particularly well suited to retrieval from a large, fixed list of canned responses. Another reason for the poor performance would be because of the weakly labelled training dataset. These results show that the contrastLSTM model was able to effectively leverage the unlabelled dataset, but the DAM network was not. Creating the negative samples within the batch during training time for the DAM model is suboptimal, as it might lead to lots of false negatives, as well as true negatives that are very easy to distinguish from true positives because they are far from the semantics of the contexts. Ideally a dedicated negative training dataset would be developed. However creating such a negative training is would be very labor intensive. Another disadvantage of the DAM model is that it is too slow to use in our CPU deployment environment, and ultimately the reason why we didn't pursue this approach any further.

\subsection{Comparison of our hierarchical architecture across 3 learning objectives}
Next we evaluate our LSTM hierarchical embedding architecture across the 3 different learning objectives described in section \ref{model-comparisons}: binary classification, multi-class classification and contrastive learning. All 3 LSTM models were evaluated on the test set of the ``true positive" dataset. As described in section \ref{Supervised-dataset}, this true positive test dataset consists of approximately 8K (context, response) pairs from the usage logs of the deployed model.
\begin{table}[h!]
  \begin{center}
    \label{LSTM-model-comparison}
    \begin{tabular}{l  | l}
      \hline 
      \textbf{Model} & True Positive Data \\
      \hline 
      contrastLSTM  & 93.42\% \\
      LSTM Binary Classifier  & 90.13\%\\
      LSTM Multi-class Classifier  & 90.12\%\\
      \hline 
    \end{tabular}
  \end{center}
  \caption{Top-3 accuracy comparison across the 3 different learning objectives.}\label{LSTM-model-comparison}
\end{table}

 As shown in table \ref{LSTM-model-comparison}, all three models achieved over 90\% accuracy for top-3 prediction on the true-positive dataset. The success of the self-supervised max-margin LSTM may stem from the fact that the self-supervised training dataset is the only one that leverages the entire raw dataset, since it includes all agent and all customer responses, with tens of thousands of unique positive targets. By comparison, for the supervised binary classification training dataset, the positive targets consist only of canned responses (of which there are only a few hundred unique responses). The other agent utterances from the raw chatlogs can still be included in the training set as negative targets, but many dialogues are discarded in the creation of this binary classification labelled dataset. The multi-class classifier discards even more of the raw data, as its training data only includes positive examples with the canned responses as targets, and there are no negative targets.

\section{Discussion} 
\subsection{Model confidence analysis}\label{Model confidence analysis}
\begin{figure}[h]
\centering
\includegraphics[width=5.3cm]{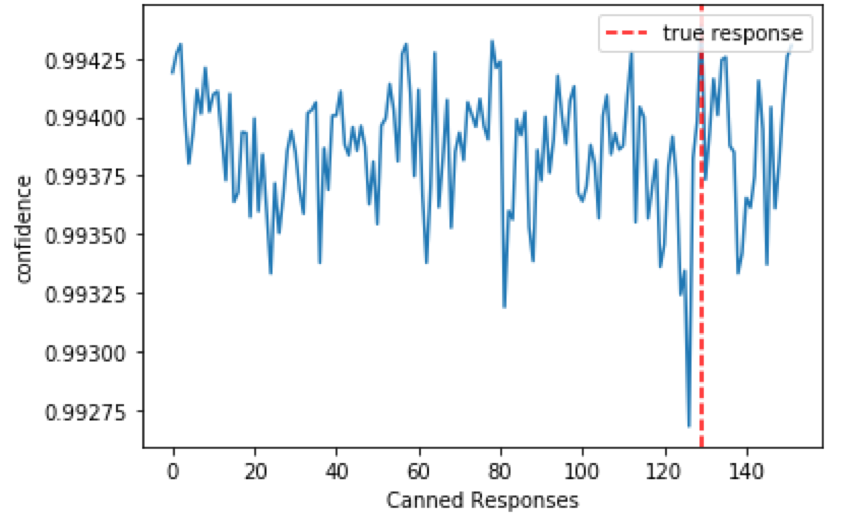} 
\caption{Model confidence (cosine similarity) across the canned responses for the contrastLSTM model, for a given example context. context}\label{max-margin-confidence}
\end{figure}

\
The task of deciding when to suggest responses (discussed in section \ref{when-to-suggest}) relies on an analysis of the model confidence. Ideally a response selection model would exhibit sparse spikes of high confidence, indicating a good ability to differentiate between appropriate and inappropriate responses. However, an analysis of the distribution of model confidence in Figure \ref{max-margin-confidence} found that the contrastLSTM model in general has relatively high confidence across the entire list of canned responses for a given context.

\begin{figure}[h]
\begin{subfigure}{4.35cm}
\includegraphics[width=4.35cm]{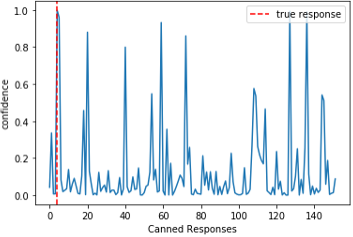}
\caption{Binary classifier}
\end{subfigure}
\begin{subfigure}{4.35cm}
\includegraphics[width=4.35cm]{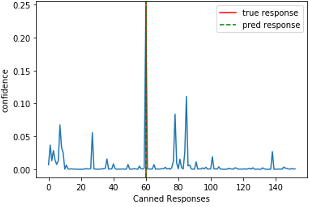} 
\caption{Multiclass classifier}
\end{subfigure}
\begin{subfigure}{4.35cm}
\includegraphics[width=4.65cm]{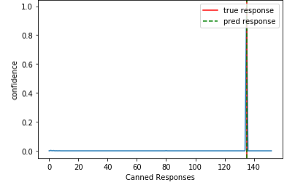} 
\caption{Multiclass classifier}
\end{subfigure}
\caption{Confidence for one example context for the binary classifier and two example contexts for the multiclass classifier.}\label{classification-confidence}
\end{figure}

It was this difficulty that originally prompted us to reformulate our LSTM model architecture as a standard classification problem, in the hope that both the binary and the multi-class classifiers of section \ref{Multiclass-classifier} would have more sparse spikes of high prediction confidences compared to the contrastive learning LSTM model. The spikes of higher confidence for the binary classifier for the example context shown in figure \ref{classification-confidence}(a) is indeed more sparse than for the cosine similarity LSTM shown in \ref{max-margin-confidence}. However, the multi-class classifier is the best by far, and was found to frequently exhibit a single dominant prediction as seen in figure \ref{classification-confidence}(b) and (c). An analysis of the confidence distribution of the multiclass classifier is presented in \ref{multi-distribution}.

\subsection{Extending the canned response list}
It is inevitable that an organisation will want to extend their canned response list over time. An advantage of self-supervised contrastive learning for chatbot deployment is that new canned responses can be added without needing to retrain the model. This is because any new canned response is likely to be similar to some existing responses that were present during training. In contrast, if the canned response list was extended for a binary classifier matching model without retraining the model, one would not expect the model to perform well on the new canned responses that weren't seen in training. The multi-class classifier matching model is the least flexible of the 3 learning objectives, as it is not possible to extend the deployed canned response list without retraining the model.

\subsection{The impact of negative sampling on the various learning approaches}
Our training approach still has room for improvement with regard to negative sampling, however. In particular, simply treating all the other samples within the batch during training as the negative samples for the max-margin loss (\ref{N-pair-loss}) is sub optimal. It might lead to false negatives, as well as true negatives that are very easy to distinguish from true positives because they are far from the semantics of the contexts. 

These are the same weaknesses we highlighted for the within-batch creation of negative samples for the binary classifier training set that the DAM model was trained on. Indeed it seems that contrastive learning and the standard binary classification approach suffer from analogous challenges with finding good negative samples. \cite{DAM2} found that developing a dedicated negative training dataset for the Ubuntu Dialogue corpus gave performance improvements over the original dataset, which was created using random negative sampling. In particular, they showed that using negative sampling strategies such as semi-hard sampling and decay-hard sampling improved performance for multiple different response retrieval classification models including DAM and the Dual-encoder LSTM.  \\

It is interesting to note that the multi-class LSTM model didn't use negative samples at all, yet has the greatest confidence its in predictions of the 3 LSTM models (as seen in Figures \ref{max-margin-confidence} and \ref{classification-confidence}). In fact, it may be the case that the random negative sampling approach contributes to the sub-optimal confidence profile of the max-margin LSTM in figure 
\ref{max-margin-confidence}, as it limits the quality of the learned embedding space. Firstly, since negative samples are pushed away from the context in the embedding space by the max-margin loss, any false negatives would be pushed away. Secondly, careful selection of negatives exposes the model to more informative negative responses that are, for example, close to the semantics of the context but still inappropriate responses for subtle reasons. A random negative sampling approach would select far fewer high quality negatives, and may instead result in a lot of very easy samples that are far from the context semantics. Interestingly though, the sophisticated negative mining approach of \cite{Wu3} gave rather modest improvements in top-1 recall rate over the random negative sampling approach.  

\subsection{Consideration of inference when constructing model architecture}
Deployment inference speed is rarely taken into account in the literature when evaluating matching models, despite being an important factor for real life chatbots. Although we experimented with more advanced model architectures, including multi-level similarity encoding and attention-based approaches, we never chose to replace our original, basic model architecture, which had the best overall performance when you also take network latency on our CPU deployment environment into consideration.

\subsection{Benefits of the contrastLSTM matching model}
Our self-supervised, contrastive learning approach presents a way to train and deploy a response retrieval chatbot directly from the raw chatlogs, without the need for creating a labelled training set. The hierarchical model architecture allows for utterance embeddings to be cached, in order to meet inference speed requirements on a CPU deployment environment. Furthermore, the self-supervised contrastive learning approach for chatbot deployment makes it is easier to add new canned responses since the model does not need to be re-trained.

The advantage of the multi-class classification model is that it is more confident in its predictions when an appropriate canned response can be used, as seen in Figure \ref{classification-confidence}. This property means it is easier to set a minimum confidence threshold when deciding whether to propose any response to the agent. If there are no appropriate response it is better to make no suggestion to the agent since inappropriate suggestions can waste an agent's time.

Table \ref{model-comparison} below shows a summary comparing the properties of the 3 different learning objectives for training a matching model for chatbot deployment.

 \begin{table}[h!]
  \begin{center}
    \label{model-comparison}
    \begin{tabular}{|l  | l | l | l|}
      \hline 
       & Contrastive & Binary & Multi-class \\
       & learning & Classification & classification \\
      \hline 
      Need labelled dataset  & No & Yes & Yes \\
      Easy to extend canned list  & Yes & No &  No \\
      Training time  & Long & Short & Short \\
      Inference performance  & Slow & Slow & Fast \\
      \hline 
    \end{tabular}
     \caption{Comparison of matching models by learning objectives.}\label{model-comparison}
  \end{center}
\end{table}

\section{Conclusion}
In this paper we provide an end-to-end framework for developing a response retrieval chatbot, starting from an unlabelled chatlog corpus. We present a deployment-centric analysis of the task of response selection, and evaluate different approaches to match a conversation context to a response candidate, including self-supervised contrastive learning, binary classification and multi-class classification.
The self-supervised LSTM model performed the best on our internal chatlog dataset. This suggests that with enough data, self-supervised response matching models have the ability to perform well, thus removing  the need to create positive/negative training datasets.

\section*{Acknowledgements} The authors are grateful to all members of the team who contributed to the development of the chatbot, and would like to give special thanks to Mark Moloney and Garry Barnes.

\bibliography{ref}

\appendix
\section{Implementation Details}
\subsection{Training dataset creation}\label{training-set}
To create the training dataset from the raw, unlabelled data, each dialogue was processed so that the ordering of utterances was uniform across the dataset (ie. the utterance order was ACACACAC..., where A = agent utterance and C = customer utterance). Each utterance was then padded with the ``$<$eos$>$" token, and then ``$<$empty$>$" tokens to make all utterances 40 words long (where any utterances longer than 40 words were truncated). A sliding window approach was then taken to create $<$context, target$>$ pairs from each dialogue, where the target was comprised of an agent utterance and the context was comprised of all utterances in the dialogue that came before the target agent utterance. In this way each dialogue from the raw data is used to create multiple $<$context, target$>$ training pairs.

\paragraph{Binary classifier training dataset}\label{binary training set}
For the binary classification models, each $<$context, target$>$ pair is given a score in $[0, 1]$ which denotes how confident the model is that the target is the correct next response. These models therefore require a training dataset like the Ubuntu Dialogue dataset, where 50\% of the $<$context, target$>$ pairs are `positive' (ie. the targets is the true next response) and 50\% `negative' (ie. the target is not a valid next response). To create an appropriate binary classification dataset from the self-supervised training dataset, 50\% of the samples in every batch were turned into negative pairs during training by shuffling their target responses. 

\subsection{Comparison models}
\paragraph{Deep Attention Matching (DAM) Network}\label{DAM classification model}
The DAM network \citep{DAM} is a supervised learning model that uses an attention module which is similar to the encoder used in the Transformer network \citep{Tran}, but extends it in two ways: (1) using stacked self-attention to harvest multi-grained semantic representations. (2) utilising cross-attention to match with dependency information across the context and response. In the representation module, $L$ identical layers of self-attention are stacked to create multi-grained representations of each context utterance $u_i$ and the target utterance $r$, denoted as $U_i$ and $R$ respectively. Then for each granularity $l\in[0...L]$, a self-attention matching matrix $M_{self}^{u_i, r, l}$ and a cross-attention matching matrix $M^{u_i,r,l}_{cross}$  are created, measuring the relevance between utterance $u_i$ and response $r$ with textual information and dependency information respectively. Those matching matrices are then merged into a 3D matching image $Q$, where 3D convolution and max-pooling is applied, followed by a linear layer to the end to make the decision whether $r$ is the true response or not.

\subsection{Additional experimental details}
For all approaches, the same vocabulary of size 5000 was used, and a skip-gram word embedding of dimension 200 was trained using fastText. 
 
\paragraph{Contrastive learning LSTM}
A vocabulary of size 5000 was used, and a skip-gram word embedding of dimension 200 was pre-trained using fastText. A batch size of 128 was used, which means the training objective was to choose the correct response from a list including the correct target and 127 negative targets. A margin $m$ of 0.0001 was used for the max-margin loss. The utterance and context embedding modules each consisted of 2 layers of LSTM with residual connections. Layer norm was applied in the utterance and target embedding modules, and a dropout keep-probability of 0.5 was applied before the linear output layer of the target and context embeddings. Overfitting was not a problem we encountered with this architecture. We investigated using different embeddings for different utterance types i.e. customer Vs agent utterances, (which was one key focus of \cite{Wang2}), however no performance improvement was observed, so this idea was abandoned. Similarly, it did not make much difference whether or not we shared weights across the context and target utterance embedding modules, so we went with shared weights. 

\paragraph{LSTM Binary Classifier}
The same word embedding, and the same sized LSTM modules as the contrastive learning model are used, except that a sigmoid activation function is applied to the concatenated context and target embeddings, and binary cross-entropy loss is used for training.

\paragraph{LSTM Multi-class Classifier}
Again the same word embedding, and the same sized stacked LSTMs are used for the utterance and context embeddings, however now the embedded context is passed to 3 linear layers. The last linear layer outputs an N-dimension tensor, (where N=290 is the number of canned responses) and then the softmax activation function is applied. 

\paragraph{DAM}
We trained 5 different models, each with a different number of attention layers between 1 and 5. To train this model we used the labelled training set of positive pairs described in section \ref{Supervised-dataset}.  In line with the standard approach, negative examples were sampled randomly, however instead of using a dedicated negative training set, as a shortcut the negative examples were sampled within the batch during training time. Like with all of our experiments, a batch size of 128 was used. We tried with and without positional encoding, but it didn't make much of a difference.

\section{Utterance embedding and similarity classification model}\label{utterance_sim}
A tool that is very handy to have for multiple facets of this project is an utterance similarity classification model. A model that can accurately classify a pair of utterances as having ``similar" intent Vs ``different" intent is useful for:
\begin{enumerate}
\item Extending the canned response list by finding popular/common agent utterances from chat logs that are dissimilar to all existing canned responses.
\item Detecting duplicates in the canned list.
\item Helping to create the negative pairs for a labelled dataset for a binary classification response prediction model.
\end{enumerate}

\noindent To train the similarity classification model, a number of approaches were compared for building the similarity classification model:
\begin{enumerate}
    \item using the Quora and MRPC similarity datasets to train a classifier on pairs of utterances embedded using the pre-trained BERT model \citep{BERT}. 
    \item using the cosine similarity of pairs of utterances embedded using a skip-thought embedding model \citep{SKIP}.
    \item using the cosine similarity of pairs of utterances embedded with the Universal Sentence Encoder \citep{USE}.
\end{enumerate}

\paragraph{Skip-thought utterance embedding model}\label{utter-embedding}
An utterance embedding model based on the skip-thought model of \cite{SKIP} was trained, with the minor difference that during training we used the current agent utterance to predict the 2 previous and 2 following utterances (instead of only 1 previous and following utterance). The encoder from the trained skip-thought model was then used as an utterance embedding model in multiple parts of this project.

\paragraph{Labelled dataset creation for similarity classification model}
To assist in the creation of this similarity classification model, we created a labelled dataset containing two types of pairs of utterance: unique and similar. For the purpose of this task, two sentences were considered similar if only one of them should be included in the canned list, otherwise, they were considered unique. Unique pairs were automatically generated by dividing each conversation into beginning, middle and end, and picking one agent utterance from each section. A pair of utterances created in this way is unique because they lie in different segments of the conversation, thus have different context. Candidates  for similar pairs were generated by choosing pairs that have a high cosine similarity between the skip-thought embeddings of the 2 utterances (exactly as in section 5.2). After similar candidates were created, manual labelling was performed. This resulted in a dataset of 3 thousand similar pairs and ten thousand unique pairs.\\

\begin{figure}
\centering
\subfloat[Skip-thought model prediction distribution]{
  \includegraphics[width=65mm]{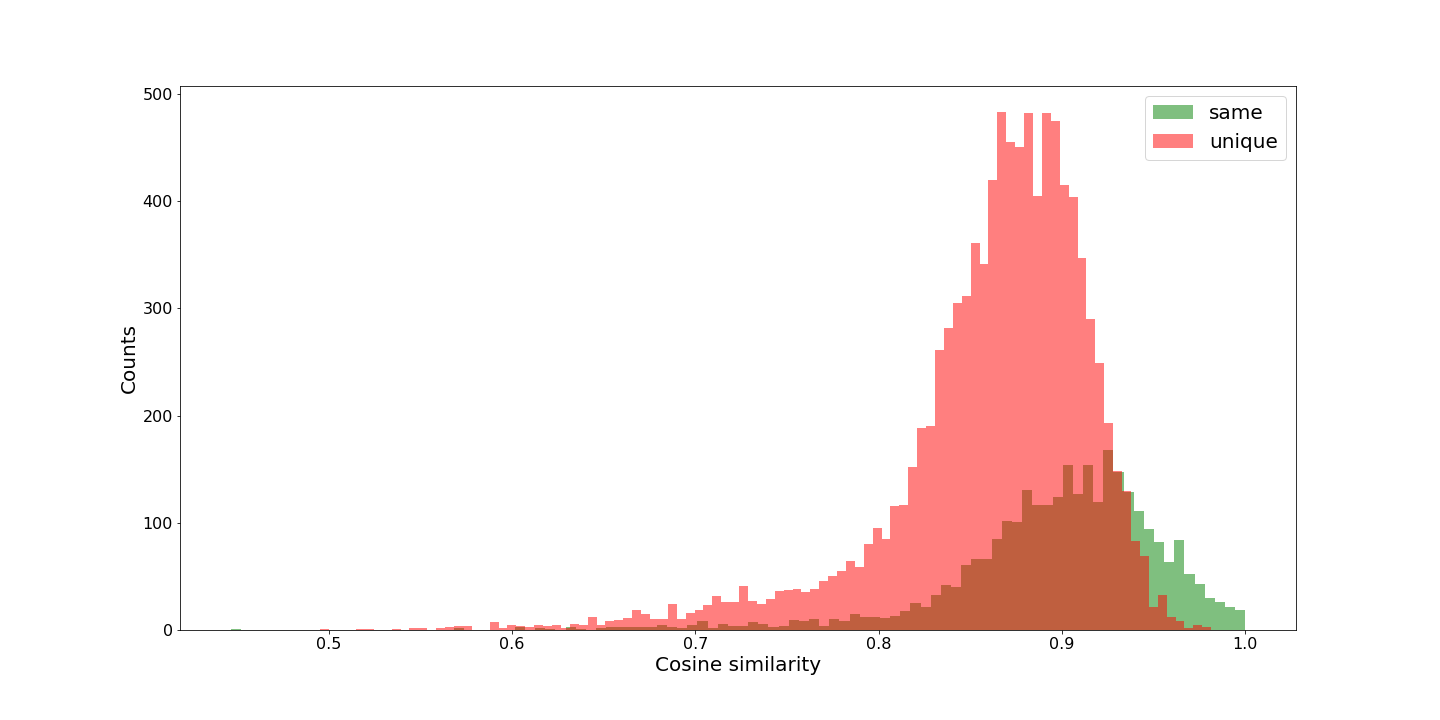}
}
\subfloat[USE model prediction distribution]{
  \includegraphics[width=65mm]{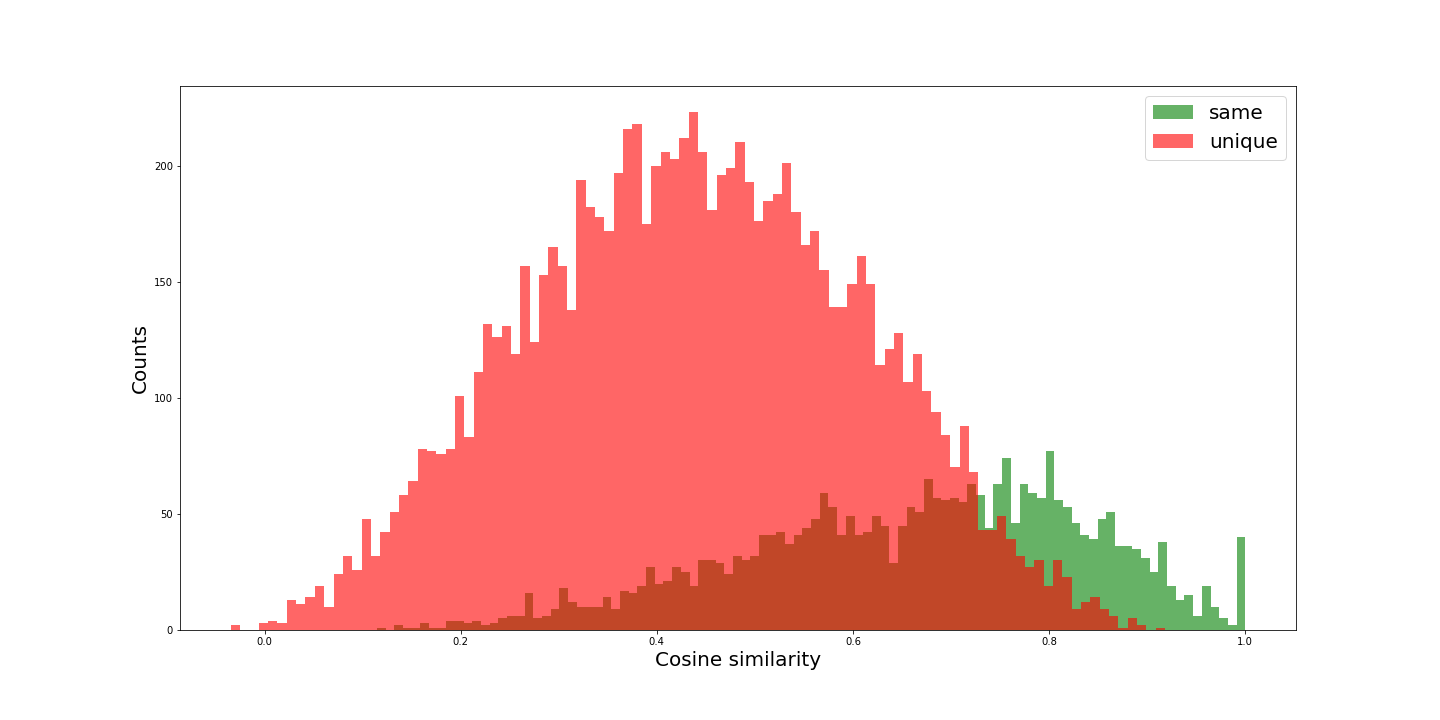}
}
\hspace{0mm}
\subfloat[BERT-MRPC model prediction distribution]{
  \includegraphics[width=65mm]{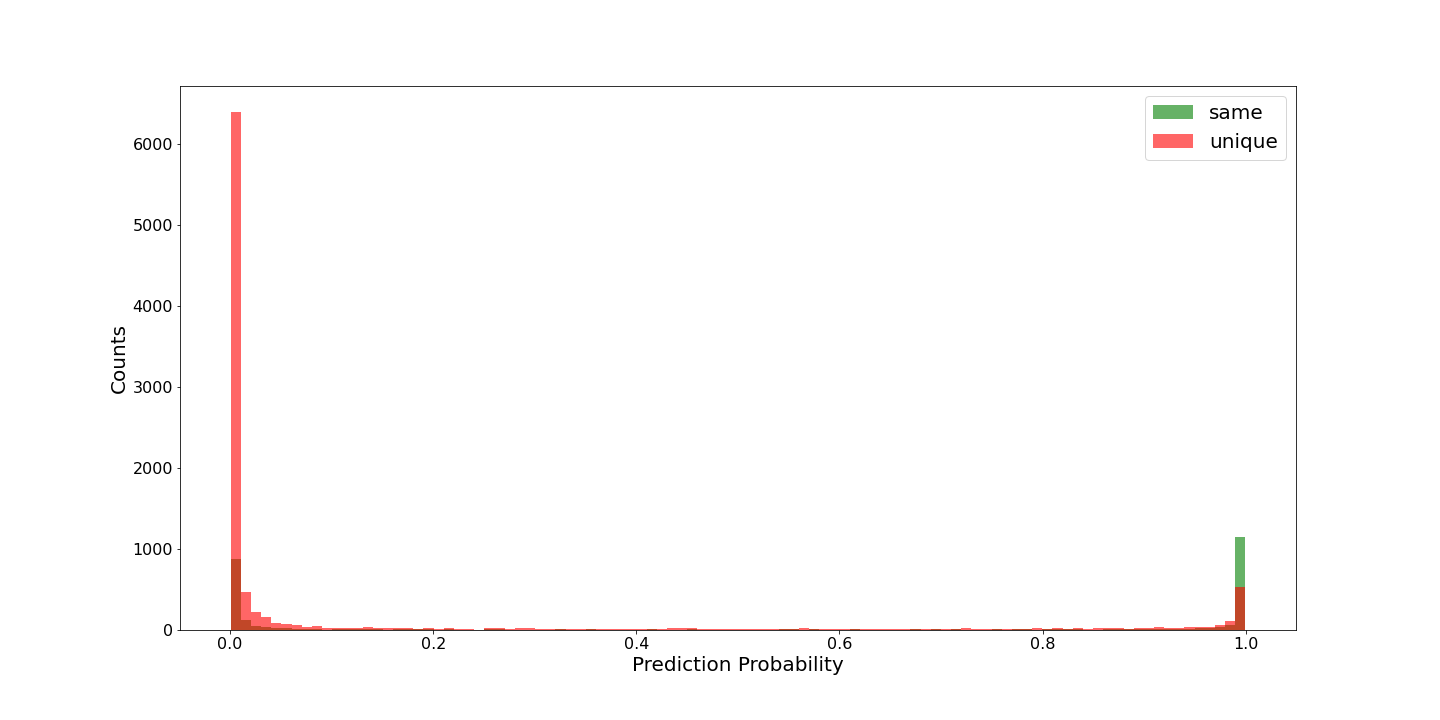}
}
\subfloat[BERT-Quora model prediction distribution]{
  \includegraphics[width=65mm]{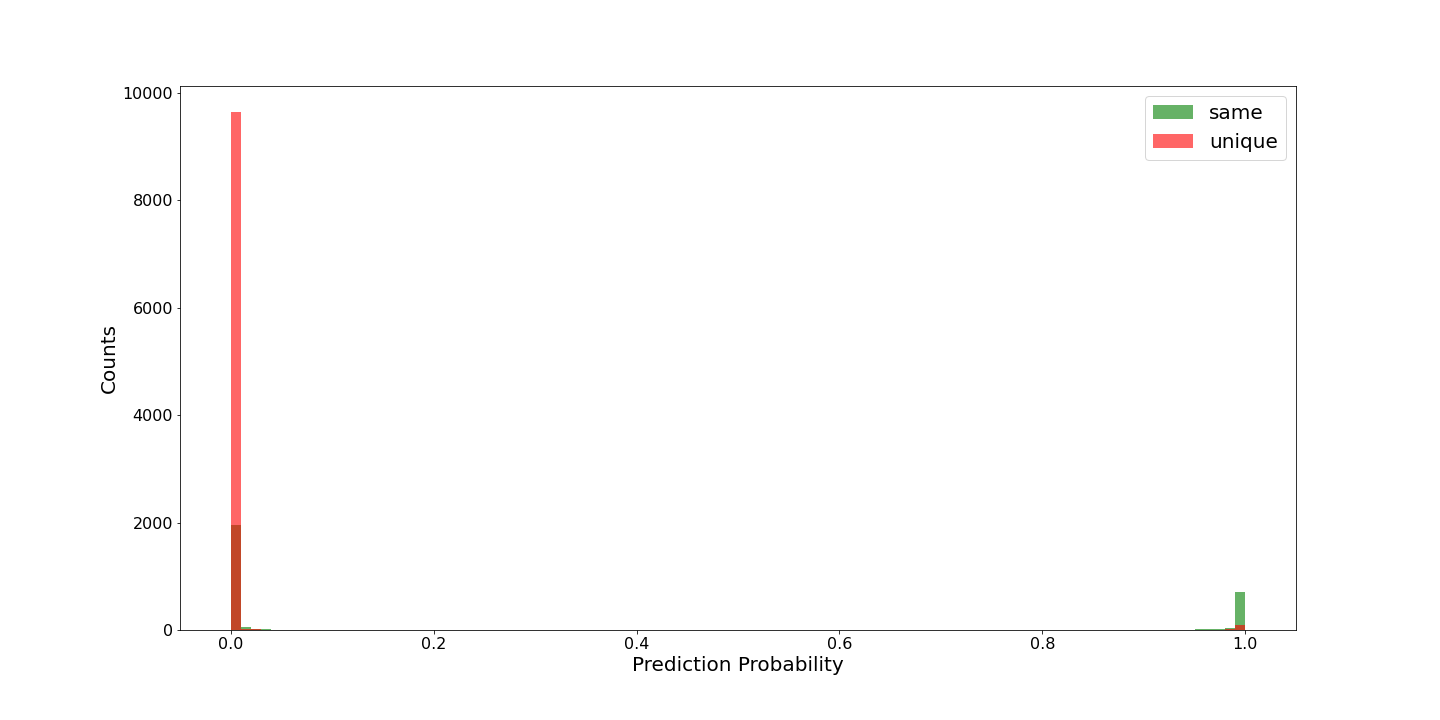}
}
\caption{Comparison of the distribution of model predictions on pairs of 'unique' and 'similar' utterances.}\label{skipthought}
\end{figure}

\paragraph{Utterance similarity classifier}
A `well' performing classifier would have a clear separation between the distributions of prediction confidences of the two classes, in order to reduce the false positive and false negative rates. In our case, the skip-thought utterance embedding from section \ref{utter-embedding} was not very good at separating the classes of this dataset based on cosine similarity, as seen in figure \ref{skipthought}(a) below, and there were not many extreme positive or negative samples. This would be partly due to the dataset, which was not perfect due to the subjective nature of human labelling, and the sampling strategy for unique pairs, which may have generated some false negatives.

A similar distribution was seen in figure \ref{skipthought}(b) when using cosine similarity with the Universal Sentence Encoder \citep{USE}. We also fine-tuned the pre-trained BERT embeddings \citep{BERT} for the similarity classification task using publicly available similarity datasets, including Quora and the Microsoft Research Paraphrase Corpus (MRPC). As expected, they scored most pairs as extreme positive or negatives, with the BERT-MRPC model in figure \ref{skipthought}(c) getting better coverage of the positive class than the BERT-Quora model in figure \ref{skipthought}(d).

To compare the above-mentioned models, we plotted the ROC curves in figure \ref{ROC-plots} and calculated the AUC for each in table \ref{AUC-table}, which is the area under the ROC curve, which can be interpreted as the probability that the scores given by a classifier will rank a randomly chosen positive instance higher than a randomly chosen negative one. Although the creators of BERT caution that it does not generate meaningful sentence vectors out-of-the-box, we measured it's performance at separating the classes of this dataset based on cosine similarity anyway, just to get a benchmark for BERT.

\begin{minipage}{\textwidth}
\begin{minipage}[b]{0.37\textwidth}
\raggedleft
\includegraphics[width=55mm]{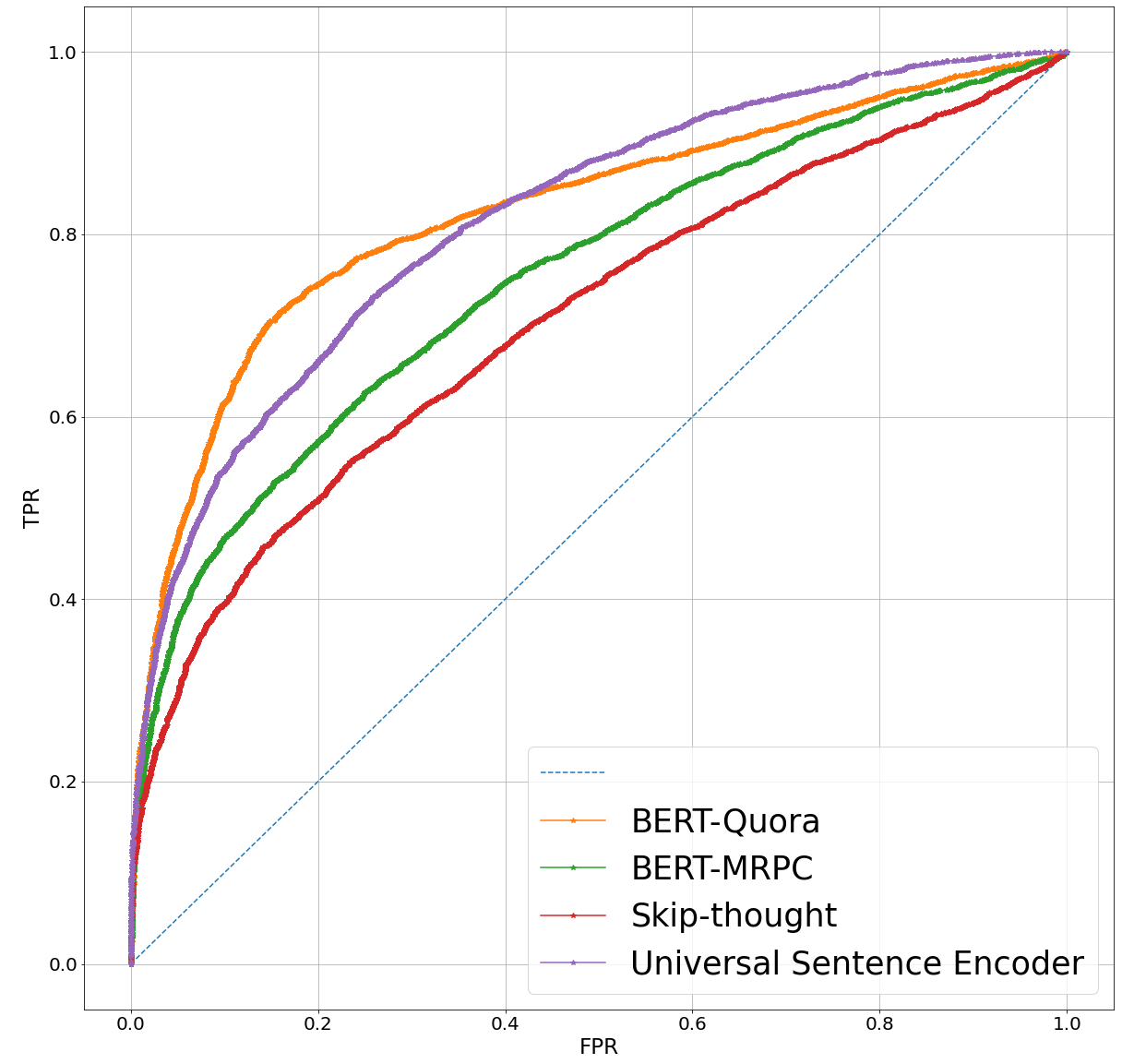} 
\captionof{figure}{ROC curves for the similarity models from figure \ref{skipthought}}\label{ROC-plots}
\end{minipage}
\qquad
\begin{minipage}[b]{0.4\textwidth}
\raggedleft
\begin{tabular}[t]{l | l}
  \hline 
  \textbf{Model} & \textbf{AUC}  \\
  \hline 
  Skip-thought & 0.72 \\
  Universal Sentence Encoder & 0.817\\
  BERT with cosine similarity & 0.69\\
  BERT + MRPC fine-tuning & 0.753\\
  BERT + Quora fine-tuning & 0.824\\
  \bottomrule 
  \end{tabular}
  \captionof{table}{AUC values for each similarity model.}\label{AUC-table}
\end{minipage}
\end{minipage}

 \subsection{Multiclass model confidence analysis.}\label{multi-distribution}
\begin{figure}[h]
\centering
\begin{subfigure}{4.3cm}
\includegraphics[width=4.3cm]{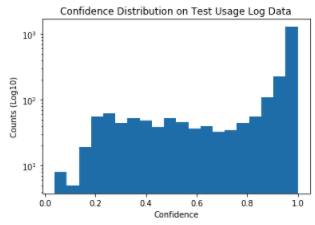} 
\caption{All predictions.}
\end{subfigure}
\begin{subfigure}{4.3cm}
\includegraphics[width=4.3cm]{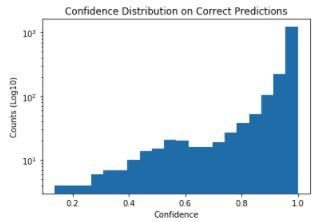} 
\caption{Correct predictions.}
\end{subfigure}
\begin{subfigure}{4.3cm}
\includegraphics[width=4.3cm]{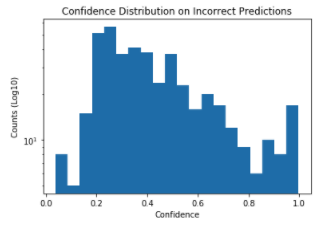} 
\caption{Incorrect predictions.}
\end{subfigure}
\caption{Multi-class classifier confidence distribution.}\label{multiclass-confidence-comparison}
\end{figure}

The multi-class model displayed the sparse prediction confidence profile we were looking for, and would also be expected to perform better in the production environment in terms of latency, as it doesn't need to compare the context with all the canned responses. This prompted us to carry out further analysis on the confidence distribution for the multi-class classification model. Using a test set of 2286 positive data points taken from the usage logs of the deployed chatbot model, the confidence distribution of correct positive and incorrect positive predictions were compared. The top-1 accuracy for the model on this dataset was found to be 81.5\%, and we see in figure \ref{multiclass-confidence-comparison} that the model is generally less confident on the examples that it gets wrong.

We also tested the model on 1000 randomly chosen negative data points and found that the confidence for negative data is usually low, see Figure \ref{multiclass-negative-confidence} below. Figures \ref{multiclass-confidence-comparison} and \ref{multiclass-negative-confidence} together can be used to help choose a confidence threshold for when the deployed model should suggest a response to the customer support agents. In general, a good choice could be when the multi-class model has a confidence of at least 0.6, or alternatively, in cases where incorrect suggestions need to be limited, a confidence threshold of 0.8 could be selected. 
\begin{figure}[h]
\includegraphics[width=5cm]{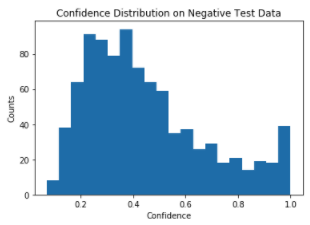} 
\caption{Multi-class classifier confidence on negative examples}\label{multiclass-negative-confidence}
\end{figure}

\end{document}